\documentclass[letterpaper, 10 pt, conference]{ieeeconf}    

\IEEEoverridecommandlockouts                                
\usepackage[pdftex]{graphicx}   
\usepackage{epsfig}     
\usepackage{amsmath}    
\usepackage{amssymb}    
\usepackage{interval}
\usepackage{textcomp}
\usepackage{gensymb}
\usepackage{graphicx}
\usepackage{mdframed}
\usepackage{subcaption} 
\usepackage{caption}
\usepackage{cite}
\usepackage{units}
\usepackage{hyperref}
\usepackage{lipsum}
\usepackage{xcolor}
\usepackage[linesnumbered,ruled]{algorithm2e}
\SetKw{KwBy}{by}
\SetKw{KwIn}{in}

\usepackage{verbatim}
\usepackage{balance}
\usepackage{acro}
\input{styles/acro.sty}
\input{styles/acro-CSE.sty}
\input{styles/acro-MMSF.sty}
\input{styles/acro-UWB.sty}

\newif\ifjournal
\journalfalse
\newif\iffinal
\finaltrue

\usepackage{amsmath}
\usepackage{amssymb}
\usepackage{mathtools}

\iffinal

    \newcommand{\removed}[1]{}
    \newcommand{\warning}[1]{}
\else

    \newcommand{\removed}[1]{\textcolor{red}{{\textit{#1}}}}
    \newcommand{\warning}[1]{\textcolor{red}{\textbf{#1}}}
\fi

\newcommand\etal{\emph{et al.}~}

\newcommand{\apri}{\emph{a-priori}~}

\newcommand{\normaldist}[2]{\mathcal{N}\left(#1,#2 \right)} 
\newcommand{\skewmat}[1]{\left[#1\right]_{{\scriptscriptstyle \!\times}}}
\newcommand{\jacobian}[3]{\left.\frac{\partial {#1}}{\partial {#2}}\right\rvert_{#3}}

\newcommand\referencediag[4]{{}^{{#2}}_{{#4}}{{#1}}^{}_{{#3}}}   
\newcommand\referencediagt[5]{{}^{{#2}}_{{#4}}{{#1}}^{#5}_{{#3}}}   
\newcommand\reference[4]{{\referencediag{#1}{#2}{#3}{#4}}}  
\newcommand\referencet[5]{{\referencediagt{#1}{#2}{#3}{#4}{#5}}}  

\newcommand\transpose{{\mathsf{T}}}

\newcommand\bZero{\boldsymbol{0}}

\newcommand\bomega{{\boldsymbol\omega}}

\newcommand\bSigma{{\mathbf{\Sigma}}}

\newcommand\bnu{{\boldsymbol\nu}}

\makeatletter
\newcommand{\generateAlphabet}[4]{%
  \def\@tempa{#1} 
  \count@=`#3
  \loop
  \begingroup\lccode`?=\count@
  \lowercase{\endgroup\@namedef{\@tempa ?}{#2{?}}}%
  \ifnum\count@<`#4
  \advance\count@\@ne
  \repeat
}

\generateAlphabet{v}{\mathbf}{A}{Z}
\generateAlphabet{v}{\mathbf}{a}{z}
\generateAlphabet{c}{\mathcal}{A}{Z}
\generateAlphabet{cc}{\mathcal}{a}{z}
\generateAlphabet{cs}{\scriptscriptstyle\mathcal}{A}{Z}
\generateAlphabet{fr}{\mathfrak}{A}{Z}
\generateAlphabet{fr}{\mathfrak}{a}{z}
\generateAlphabet{rm}{\mathrm}{A}{Z}
\generateAlphabet{rm}{\mathrm}{a}{z}

\newcommand\Sensor{{\mathsf{S}}}

\newcommand\Scenario{{\mathrm{S}}}
\newcommand\Experiment{{\mathrm{E}}}

\newcommand\HistBelief{\boldsymbol{\cX}}
\newcommand\HistMeas{\boldsymbol{\cZ}}
\newcommand\HistSigma{\boldsymbol{\cC}}

\newcommand\ID{\mathsf{id}}

\newcommand\BufferDict{\mathsf{Dict}}
\newcommand\BufferHist{\mathsf{Hist}}

\newcommand\InstanceHandler{{\mathsf{H}}}

\newcommand\state{\vx}

\newcommand\stateest{\hat{\vx}}
\newcommand\stateerr{\tilde{\vx}}

\renewcommand\reference[4]{{\referencediag{#1}{#2}{#3}{#4}}}  

\newcommand\REMOVED[1]{{}}

\newcommand{\removelatexerror}{\let\@latex@error\@gobble}

\usepackage[noabbrev,capitalise]{cleveref} 

\title{{\LARGE \textbf{Modular Meshed Ultra-Wideband Aided Inertial Navigation with Robust Anchor Calibration}}}

\author{Roland~Jung$^{1}$, Luca~Santoro$^{2}$, Davide~Brunelli$^{2}$, Daniele~Fontanelli$^{2}$, and~Stephan~Weiss$^{1}$
\thanks{$^{1}$Roland Jung and Stephan Weiss
        are with the Department of Smart Systems Technologies
        in the Control of Networked Systems Group,
        University of Klagenfurt, Austria
        {\tt\small \{first.last\}@ieee.org} }
\thanks{$^{2}$Luca Santoro, Davide Brunelli and Daniele Fontanelli are with the
Department of Industrial Engineering, University of Trento, Italy
        {\tt\small \{first.last\}@unitn.it} }
\thanks{This work was supported by the Austrian Ministry of Climate Action and Energy (BMK), grant agreement 891124 (RoMInG) \& 895156 (SwarmNav).}
}

\begin{document}
\maketitle
\thispagestyle{empty}
\pagestyle{empty}

\begin{abstract}\label{sec:abstract} 
{This paper introduces a generic filter-based state estimation framework that supports two state-decoupling strategies based on cross-covariance factorization. 
These strategies reduce the computational complexity and inherently support true modularity -- a perquisite for handling and processing meshed range measurements among a time-varying set of devices.
In order to utilize these measurements in the estimation framework, positions of newly detected stationary devices (anchors) and the pairwise biases between the ranging devices are required. In this work an autonomous calibration procedure for new anchors is presented, that utilizes range measurements from multiple tags as well as already known anchors. To improve the robustness, an outlier rejection method is introduced. After the calibration is performed, the sensor fusion framework obtains initial beliefs of the anchor positions and dictionaries of pairwise biases, in order to fuse range measurements obtained from new anchors tightly-coupled. The effectiveness of the filter and calibration framework has been validated through evaluations on a recorded dataset and real-world experiments.}
 
\end{abstract}
\vspace{-2mm}
\section{Introduction}
\label{sec:introduction}
{
Autonomous robots have seen a steady progress and obtained significant interest. With many applications across diverse fields, such as the exploration of hazardous environments or infrastructure inspection, the requirement for an accurate, cost-effective and scalable localization technology is on the rise, as it forms the pillar for feedback control and path planning. 
In particular, operating in \ac{GPS}-denied environments, such as (partially) structured indoor environments, turns into a challenge. 

These scenarios often require different reliable localization methods, such as Visual-, LiDAR-, or Radar-Inertial Odometry~\cite{geneva_openvins_2020, xu_fast-lio_2021,michalczyk_multi_2023}. However, these approaches suffer from accumulated drift over time and are computationally demanding. In addition, vision-based localization methods are prone to fail under challenging visual conditions, while LiDAR sensors are typically heavy and expensive. 
Alternatively, a range-based localization infrastructure can be used\REMOVED{, e.g., based on \ac{TOA}, \ac{TDOA}, \ac{RTOF}~\cite{neuhold_hipr_2019}, or \ac{RSS} measurements to estimate inter-antenna distances between transceivers~\cite{sahinoglu2008ultra}}. 
A promising technology for both data transmission and precise localization are the \ac{UWB} radio frequency signals~\cite{sahinoglu2008ultra}. 
For exploration tasks, the localization coverage is limited by the communication range and can be increased by deploying additional stationary \ac{UWB} anchors in the area of interest as shown in~\cref{fig:MMSF-sensor-constellation}. 
%
\begin{figure}[t]
  \centering 
  \includegraphics[trim={0 0 0 0},clip,width=\linewidth]{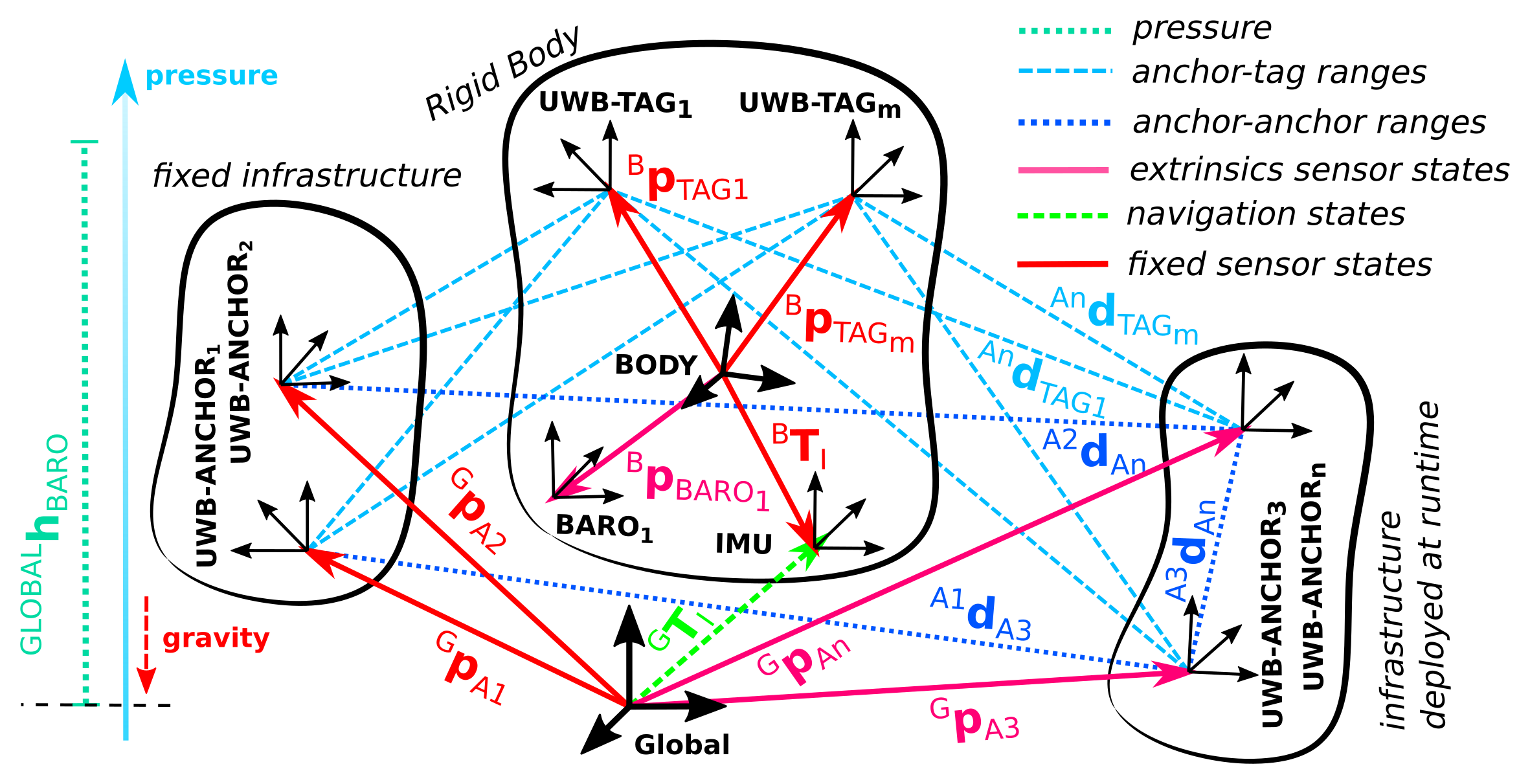}%
  \caption{{Spatial frame constellation of the proposed modular aided inertial navigation framework. A calibrated \ac{UWB} ranging infrastructure is extended at runtime by additional stationary ranging devices. After a coarse initialization, these new anchors positions (purple) are estimated in filter framework.}}\label{fig:MMSF-sensor-constellation}
  \vspace{-7mm}
\end{figure}
%
The uncertainty experienced during exploration is related to the sensors employed and the placement of additional anchors sensors. Thus, optimal anchor positions can be found~\cite{bulusu2001adaptive,moreno2016optimal,magnago2019effective,santoro2022online}. After placement, the anchor position can be estimated by solving an optimization problem~\cite{hol_ultra-wideband_2010,hausman_self-calibrating_2016,delama2023uvio}, while an optimal path can be generated in order to reduce the \ac{GDOP} of anchors and the mapping uncertainty~\cite{blueml_bias_2021,delama2023uvio}.
This initial anchor position estimate can be further refined by tightly fusing range measurements obtained by these anchors in the filter formulation, leading to an online self-calibration. Furthermore, the self-calibration can be improved by incorporating also measurements between anchors~\cite{jung_scalable_2022} or other sensor modalities, and motivates us to utilize a fully-meshed ranging scheme.

In our case, a single range measurement between two devices is estimated by a symmetric double-sided \acl{TWR} (SDS-TWR). The SDS-TWR protocol~\cite{sang2018analytical} mitigates the clock-skew, but, in order to improve the ranging accuracy, the antenna delays need to be calibrated. Thus, we model the antenna delay and other error components as pairwise biases between devices. To recover the pairwise biases, between the \apri known and deployed devices, we calibrate them in a unified multi-tag to anchor initialization routine. 
Consequently, the dimension of the state vector increases with the number of added anchors, while at the same time, the total number of measurements processed is increasing -- a known issue in EKF-SLAM approaches~\cite{thrun_simultaneous_2004}. 

To mitigate that issue, this paper introduces a generic filter-based state estimation framework that supports two state-decoupling strategies, based on cross-covariance factorization~\cite{roumeliotis_distributed_2002}. 
These decoupling methods, inherently permits true modularity, in the sense that new sensor related states (extrinsic or intrinsic) can be added or removed at runtime, which is a prerequisite for sensor hot-swapping.
To address the scalability issue, one strategy is approximating the cross-covariances to reduce the number of involved states in the filter update step and renders the approach almost invariant to the total number of sensors used.

The efficacy and the true modularity of the proposed framework is demonstrated by a fly-by anchor position and bias-calibration in offline evaluations using real-world data and in closed-loop flights. \REMOVED{In these evaluation, we could underline the advantage of the proposed unified any-device to anchor over a classical tag to anchor initialization algorithm and utilizing fully meshed range measurements to improve calibration of runtime deployed anchors.}}

\section{Related Work}
\label{sec:rw}

In the past decades, several \ac{UWB}-ranging based localization approaches have been presented, with classical approaches that are tightly fusing inertial and range information~\cite{hol_tightly_2009,hausman_self-calibrating_2016}, over to drift compensation for \ac{VIO} approaches~\cite{nguyen_range-focused_2021,delama2023uvio}, and swarm localization~\cite{xu_decentralized_2020,xu_omni-swarm_2022,shalaby_relative_2021}. 
This section briefly investigates on related aided inertial fusion frameworks and anchor calibration approaches.

In~\cite{jung_scalable_2022}, a Monte Carlo simulation to evaluate both the filter consistency and accuracy using different filter decoupling strategies and sensor constellations was conducted. The feasibility of calibrating anchor position estimates through meshed range measurements was shown. At the same time, the computational load in the estimation framework was reduced by utilizing state decoupling techniques.

Recently, Delama \etal proposed UVIO in~\cite{delama2023uvio}, consisting of an open-source \ac{MSCKF} framework combing \ac{UWB} ranging with \ac{VIO} allowing for robust and low-drift localization and a calibration framework. The calibration tool is utilizing a novel least squares formulation to solve both, the anchor positions and biases between tuples. 

In this work, we use the calibration tool, while instead of relying on camera information to estimate the 6-DoF pose of the tags during the calibration routine, we fuse range measurements tightly-coupled from a set of \apri calibrated anchors in a modular estimation framework that supports \ac{EKF} decoupling strategies approaches proposed in~\cite{jung_modular_2021}. A similar approach was proposed in~\cite{shi_anchor_2019}, however, it may face limitations in scalability, is not considering biases in the calibration routine, and was evaluated in simulations, only.

The calibration routine utilized for UVIO supports unidirectional ranging from one tag to a set of unknown anchors, only. Therefore, we reformulate the problem definition to support measurement from multiple tags and a set of known anchors to a set of unknown anchors. Since, we are aiming at meshed bidirectional ranging among \ac{UWB} devices, we need to account for the pairwise biases between these devices, and adapted the calibration routine accordingly. As the anchor calibration is sensitive to measurement noise and outliers, we integrated an outlier detection method based on \ac{RANSAC}~\cite{fischler1981Ransac}. 

Our approach improves and extends the calibration routine of~\cite{delama2023uvio} and extends the evaluations conducted in~\cite{jung_scalable_2022} with real data and closed-loop experiments.The key contributions can be summarized as follows:
\begin{itemize} 
  \item We implemented a real-time capable and truly modular multi sensor fusion framework.
  \item We created a meshed ranging dataset~\cite{jung_2024_uwbdataset} and made our range evaluation tool available online.
  \item We extended the calibration framework of~\cite{delama2023uvio} to support meshed anchor position and pairwise bias calibration, and integrated an outlier rejection method. 
  \item We conduct offline evaluations on the recorded dataset and closed-loop experiments to demonstrate the efficiency of the proposed framework.
\end{itemize}

\section{Methodology}
\label{sec:Problem-Formulation}

{In this section, we provide an overview of the proposed system. A \ac{UAV} is equipped with a 6-DoF \ac{IMU}, a barometric pressure sensor, and multiple \ac{UWB} ranging devices rigidly attached to the body frame as illustrated in~\cref{fig:MMSF-sensor-constellation}. Range measurements to a set of stationary and \apri calibrated \ac{UWB} ranging devices (anchors) are used to estimate the \ac{UAV}'s pose to perform missions autonomously. Once new anchors are detected range measurements between them and the known ranging devices together with estimated \ac{UAV} pose are collected in the calibration tool. Once sufficient measurements are received, the calibration is triggered and the initial anchor position and pairwise bias estimates are provided to the estimation framework. These fly-by initialized anchor estimates are continuously refinement and allows for extending the ranging area.

\subsection{Modular Aided-Inertial Filter Design}
\label{subsec:Modular-Aided-Inertial-Filter-Design}

At the core of the proposed estimation framework is the \textit{Instance Handler} $\InstanceHandler$, as shown in~\cref{fig:MMSF-block-diagram}, to unify $N$ held isolated filter instances (nodes) $\Sensor_{i}$, which can be registered and removed at run-time.
\begin{figure}
  \centering 
  \includegraphics[trim={5 5 40 0},clip,width=0.9\linewidth]{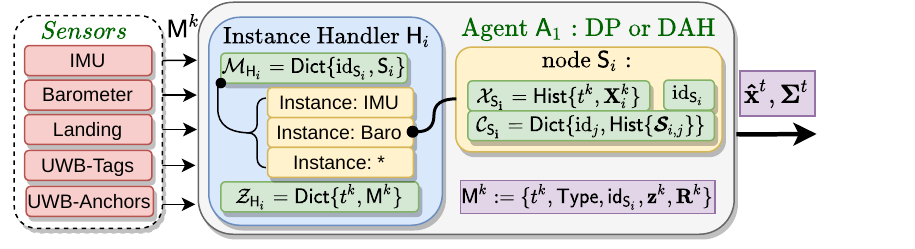}%
  \caption{Block diagram of the proposed sensor fusion framework utilizing the \acs{DAH} or \acs{DP} approach proposed in~\cite{jung_modular_2021}, consisting of an instance handler $\InstanceHandler$ maintaining instances (nodes) of a specific type.}
  \label{fig:MMSF-block-diagram}
  \vspace{-6mm}
\end{figure}
The set of heterogeneous sensors, each associated to an instance $\Sensor_{i}$,  provide measurements to the $\InstanceHandler$ through their individual callbacks. In this context, isolated means that the state variables $\vX_{i} \sim \normaldist{\hat{\vx_i}}{\bSigma_{ii}}$ of individual instances are maintained decoupled. Once needed, the state variables are concatenated, processed, and split up again. The major challenge originates from correlations $\bSigma_{i,j}$ between the variables, which need to be reconstructed accordingly and considered in each filter step~\cite{jung_modular_2021}. Each sensor needs a unique identifier and belongs to a known class of sensor types, which defines the estimated parameter/state space, constant parameters, and the sensor's dynamic model, and a method to process sensor-specific measurements. Further, each sensor is associated to an instance (node), maintaining a history of data, a unique identifier $\ID_{\Sensor}$ and, in case of observations, access to other instances' estimates through the handler. More precisely, two buffers are maintained per instance:
$\HistBelief_{\Sensor_{i}} := \mathsf{Hist} \{ t^{k}, \mathbf{X}_{i}^{k}\} $ for the beliefs and $\HistSigma_{\Sensor_{i}} := \BufferDict \{ \text{id}_j, \mathsf{Hist} \{ \mathbf{\mathcal{S}}_{i,j}\} \} $ for the factorized cross-covariances $\mathcal{S}$, according to $\bSigma_{i,j} = \mathcal{S}_{i,j} \mathcal{S}_{j,i}^\transpose$. The instance handler maintains a dictionary of instances and a history $\HistMeas_{\InstanceHandler} := \BufferHist \{ t^{k}, \mathbf{\mathsf{M}}^{k}\} $ of the processed measurements $\mathsf{M}$, which contain the measurement $\vz$, measurement noise $\vR$, $\ID_{\Sensor}$, and the timestamp $t^k$ to handle out-of-sequence measurements. $\mathsf{Hist}$ is a sorted sliding window buffer.

Currently, the handler $\InstanceHandler$ supports two state decoupling and fusion strategies~\cite{jung_modular_2021}: \ac{DP} and \ac{DAH}. \ac{DP} is a centralized-equivalent approach allowing isolated state prediction steps, i.e., this steps can be performed independently, but performs update steps on the full state. \ac{DAH} performs update steps isolated among directly involved sensor instances, reducing the computation cost at the cost of approximations made. These strategies inherently support a time-varying state space, and thus, true modularity w.r.t the sensor constellation.}

In the rest of this subsection, we summarize~\cite{jung_scalable_2022}, providing an overview of the measurement models and state definitions of the used sensor instances $\Sensor$. The sensor suite consists of a varying number of stationary \ac{UWB} anchors and \ac{UWB} tags, a barometer, and an \ac{IMU} as state prediction sensor. In contrast to the prior work, a zero velocity update is incorporated in the \ac{IMU} instance as private isolated updates.

\subsubsection{Inertial Measurement Unit}
\label{subsub:IMU-sensor}

We employ the Inertial Measurement Unit (IMU) as the sensor for state propagation within an indirect filter formulation~\cite{weiss_real-time_2011} and  with following state definition
\begin{equation}
\label{eq:IMU-state}
    \state_{\cI} = \left[ \reference{\vp}{\cG}{\cI}{}, \reference{\vv}{\cG}{\cI}{}, \reference{\vq}{\cG}{\cI}{}, \reference{\vb_{\omega}}{\cI}{}{}, \reference{\vb_a}{\cI}{}{}  \right]^\transpose
\end{equation} 
with $\reference{\vp}{\cG}{\cI}{}$,
$\reference{\vv}{\cG}{\cI}{}$, and $\reference{\vq}{\cG}{\cI}{}$ being the position, velocity, and orientation of the \ac{IMU}
$\cI$ with respect to the global navigation frame $\cG$, and
$\reference{\vb_{\omega}}{\cI}{}{}$ and $\reference{\vb_a}{\cI}{}{}$
as the estimated biases of the gyroscope and accelerometer
readings, respectively. The kinematics and measurements are described, e.g., in~\cite{jung_scalable_2022}.

\subsubsection{Barometric altimeter}\label{subsub:pressure-sensor}

Pressure readings, defined as 
\begin{equation}\label{eq:baro-measurement}
    \reference{z}{\cG}{\cP}{} = \reference{P}{\cG}{\cP}{} + \nu_{\cP},
\end{equation}
with a Gaussian noise $\nu_{\cP} \sim \cN\left(0, \sigma_{\cP}^{2}\right)$, are converted into relative height measurements given a reference pressures using the 
a standard barometric height model~\cite{bevermeier_barometric_2010}. 
The estimated barometric height, as represented by the state variables, is given by
\begin{align}
\label{eq:barometric-height}
  \reference{h}{\cG}{\cP}{} &= \left(\reference{\vp}{\cG}{\cI}{} + \reference{\vR}{\cG}{\cI}{} \reference{\vp}{\cI}{\cP}{} \right)_{z}~,
\end{align}
and is fused by defined a residual between the estimated and measured height.

\subsubsection{Range sensor}
\label{subsub:uwb-range-sensor}

The SDS-TWR scheme mitigate clock-skew-dependent bias, but there are other sources for inaccuracies, such as the antenna delay caused by the communication delays between the microchip and the \ac{UWB} antenna or the relative antenna pose~\cite{Shalaby2023Calibration}. 
An analytic error model for SDS-TWR according to~\cite{sang2018analytical} is
\begin{equation}
    \hat{t}_{\text{TOF}} - {t}_{\text{TOF}} \approx \frac{1}{4}(\xi_{ABA} - \xi_{BAB}) t_{\text{reply}},
\end{equation}
with round-trip time delay $\xi_{ABA}$ and $\xi_{BAB}$, and assuming a common reply time $ t_{\text{reply}}$ for both devices.
To compensate the ranging errors between devices, the symmetric double-sided two-way ranging (SDS-TWR) measurement between a device $k$ and $l$, is modeled as
\begin{equation}
\label{eq:UWB-meas-range}
    \reference{z}{k}{l}{} = \beta_{k,l} \reference{d}{k}{l}{}+ \gamma_{k,l}+ \nu_{\text{d}_{k}},
\end{equation}
with a pairwise and {symmetric} constant bias $\gamma_{k,l} = \gamma_{l,k}$, a symmetric distance-dependent bias $\beta_{k,l} = \beta_{l,k}$, and a  noise $\nu_{d_{k}}\sim \cN\left(0, \sigma_{d_{k}}^{2}\right)$. The distance for the individual tuples is
\begin{align}\label{eq:meas-est-uwb-range1}
  \reference{d}{\cT}{\cA}{} &= \reference{d}{\cA}{\cT}{}  = {\big\|} \reference{\vp}{\cG}{\cA}{} - \reference{\vp}{\cG}{\cT}{} {\big\|}_{2}, \\ \label{eq:meas-est-uwb-range2}
  \reference{d}{\cA_i}{\cA_j}{} &= {\big\|} \reference{\vp}{\cG}{\cA_j}{} - \reference{\vp}{\cG}{\cA_i}{} {\big\|}_{2}~,
\end{align}
where $\cT$ and  $\cA_i$ are the tag and anchor indices.

\subsubsection{Zero Velocity Update}

A detected standing still, e.g., by the flight controller, implies zero body velocity, acceleration and angular velocity~\cite{hartley_contact-aided_2018,geneva_openvins_2020}.
Once a stand-still condition is detected, we induce
noisy \emph{pseudo} observations in the form
\begin{equation}
    \vz_{\bZero} = \begin{bmatrix} \reference{\va}{\cI}{}{} + \bnu_{\va} = \bZero \\  \reference{\bomega}{\cI}{}{} + \bnu_{\bomega} = \bZero \end{bmatrix} = h_{\bZero}(\state) + \bnu,
\end{equation}
with a measurement noise of the accelerometer $ \bnu_{\va}$ and gyroscope $ \bnu_{\bomega}$. The zero velocity detection is associated with the latest \ac{IMU} readings. From the \ac{IMU} measurement model, with the gravity vector $\reference{\vg}{}{}{\cG}$, we obtain
\begin{subequations}\label{eq:fund-abs-pose-measusurement-model}
    \begin{equation*}
    \reference{\va}{\cI}{}{} = \bZero =  \referencet{\va}{\cI}{}{}{\#}  - (\reference{\vR}{\cG}{\cI}{})^\transpose \reference{\vg}{\cG}{}{}  - \reference{\vb_{\va}}{\cI}{}{} - \bnu_{\va},
    \end{equation*}
    \begin{equation*}
    \reference{\bomega}{\cI}{}{} = \bZero = \referencet{\bomega}{\cI}{}{}{\#} - \reference{\vb_\bomega}{\cI}{}{} - \bnu_{\bomega}~.
    \end{equation*}
\end{subequations}
We can linearize the measurement error $\tilde{\vz}$ around the current estimate $\stateest$ to obtain the Jacobian $\vH = \begin{bmatrix}
\vH_{\va} \\ \vH_{\bomega}
\end{bmatrix} = \jacobian{\tilde{\vz}}{\state}{\stateest} \jacobian{\state}{\stateerr}{}$, leading to following the derivatives with respect to the error-states ($\stateerr = \stateest^{-1} \oplus \state$)
\begin{equation}
    \jacobian{\tilde{\vz}_{\va}}{\reference{\tilde{\bomega}}{\cI}{}{}}{\stateest} = -\skewmat{\referencet{\hat{\vR}}{\cG}{\cI}{}{\transpose} \reference{\vg}{\cG}{}{}}, 
    \jacobian{\tilde{\vz}_{\va}}{\tilde{\reference{\vb_\va}{\cI}{}{}}}{\stateest} =     \jacobian{\tilde{\vz}_{\bomega}}{\tilde{\reference{\vb_\bomega}{\cI}{}{}}}{\stateest} = -\vI.
\end{equation}

\subsection{Fully Meshed Ranging Scheme}
\label{subsec:twr-scheme}

{Since \ac{UWB} ranging is based on broadcasting messages on the same frequency channel, meshed ranging among all nodes requires a technique to avoid collision. We propose a static time-division multiple access scheme, dividing the fully-meshed ranging cycle into $N(N-1)$ SDS-TWR cycles with a fixed timeout of $\Delta t$ as shown in~\cref{fig:scheduling}. Each ranging node has a unique ID and knows the total number of ranging devices that are participating in the meshed cycle. Each node listens to broadcasted messages and waits until the device with the lower ID has completed its $N-1$ SDS-TWR cycles. 
%
\begin{figure}[t]
  \centering 
  \includegraphics[trim={0 5 0 40},clip,width=0.8\columnwidth]{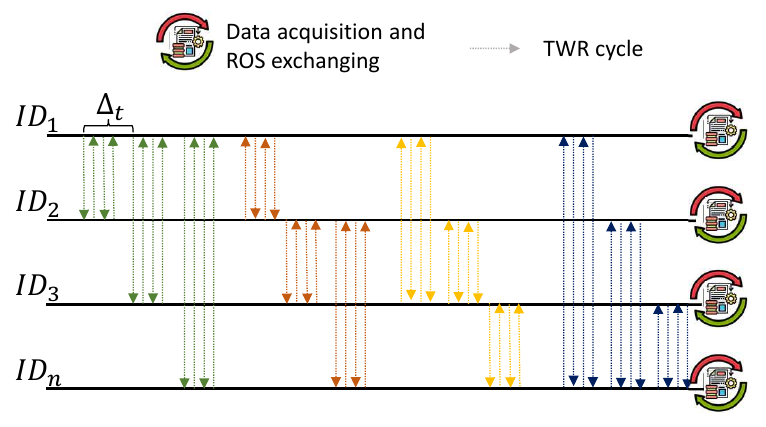}
  \caption{Congestion free fully-meshed scheduling protocol. Each node listens and waits for it's preceding node to finish the one-to-all ranging cycle. Four arrows indicate a full SDS-TWR cycle consisting of four messages, which needs to be completed within $\Delta t$.}
  \label{fig:scheduling}
  \vspace{-4mm}
\end{figure}
%
The resulting cycle rate $f_{\text{mesh}}$ for the fully-meshed ranging exhibits a quadratic relationship with $N$ nodes 
\begin{equation}
  \label{eq:update_rate}
  f_{\text{mesh}} = \frac{1}{\Delta t((N)-1)^2}~,
\end{equation}

where $\Delta t$ is a predefined time slot within a SDS-TWR cycle needs to be completed. 
Please note that this scheduling approach is sensitive to message drops and corruption, and is not dynamically adapting to a changing network configuration, but is sufficient for the experiments conducted.}

\subsection{Meshed UWB Anchor Initialization}
\label{subsec:Meshed-UWB-Anchor-Inii}

In order to be able to exploit fully meshed \ac{UWB} ranges to and from runtime deployed anchors in the filter, an initial position and the pairwise biases are requires. This is achieved by an initialization module, that requires an estimate of the robot's pose, the known positions of the UWB tags on the robots and the known positions of the already known (pre-calibrated) UWB anchors. Together, with collected range measurements to newly deployed anchors from a short trajectory, an optimization problem is solved. The solution is provided to the estimation framework to create new filter instances. In the following sections we extend the problem formulation of~\cite{delama2023uvio} to support multiple UWB tags and pre-calibrated UWB anchors.

\subsubsection{Coarse anchor positioning: Optimal Double Method}\label{sec:problem-initialization-double-method}

The optimal double method of~\cite{delama2023uvio} overcomes poor solutions obtained by standard least squares problem~\cite{hausman_self-calibrating_2016} by increasing the condition number of the least squares problem and minimizing the uncertainty of the solution.
Using a meshed ranging scheme and ignoring for now the distance related bias $\beta_{i,x} = 1$, we obtain the following measurements to and from an anchor $\cA_x$
\begin{equation}\label{eq:meas-uwb-range}
    z = || \reference{\vp}{\cG}{\cU_i}{} -  \reference{\vp}{\cG}{\cA_x}{} || + \gamma_{i,x} + \nu_d, 
\end{equation}

with $\nu_d \sim \normaldist{0}{\sigma_{d}^{2}}$ and $\reference{\vp}{\cG}{\cU_i}{} = \{\reference{\vp}{\cG}{\cT_i}{}, \reference{\vp}{\cG}{\cA_i}{}\}$ referring to tag or other anchor positions. By ignoring the measurement noise $\nu_d$ and squaring ~\cref{eq:meas-uwb-range}, a linear relation between $\reference{\vp}{\cG}{\cA_x}{}$ and $\gamma_{i,x}$ can be derived considering two distinct range measurements involving the same anchor $\cA_x$ and the same reference device at different time steps $\{t_1, t_2\}$  $\referencet{\vp}{\cG}{\cU_i}{}{t_1}$ and $\referencet{\vp}{\cG}{\cU_i}{}{t_2}$, we get
\begin{equation}
\begin{split}
    (z^{t_1})^2 - (z^{t_2})^2 =& || \referencet{\vp}{\cG}{\cU_i}{}{t_1} ||^2 - || \referencet{\vp}{\cG}{\cU_i}{}{t_2} ||^2 + 2(z^{t_1}-z^{t_2})\gamma_{i,x}\\
     & - 2\left(\referencet{\vp}{\cG}{\cU_i}{}{t_1} - \referencet{\vp}{\cG}{\cU_i}{}{t_2}  \right)^\transpose \reference{\vp}{\cG}{\cA_x}{}.
\end{split}
\end{equation}

As outlined in~\cite{delama2023uvio}, the optimal measurement pair $z^{t_1}$ and $z^{t_2}$ can be found by minimizing the uncertainty of the solution.  
The linear relation finally leads to a following least squares formulation $\vA\vx = \vb$ with $\vA = \left[\cdots; \vA_{k,i}; \cdots\right]$, $~\vb = \left[\cdots; \vb_{i}; \cdots\right]$ and
\begin{subequations}
  \begin{equation}
    \vA_{k,i} := \left(\referencet{\vp}{\cG}{\cU_i}{}{t_1} - \referencet{\vp}{\cG}{\cU_i}{}{t_2}  \right)^\transpose \left( z_{i}^{t_1} - z_{i}^{t_2} \right)
  \end{equation}   
  \begin{equation}
      b_{k} := \frac{1}{2} \left(\left( (z_{i}^{t_1})^2 - (z_{i}^{t_2})^2 \right) - || \referencet{\vp}{\cG}{\cU_i}{}{t_1} ||^2 - || \referencet{\vp}{\cG}{\cU_i}{}{t_2} ||^2 \right)
  \end{equation}
  \begin{equation}
      \vx := \begin{bmatrix} \reference{\vp}{\cG}{\cA_x}{} &  \boldsymbol{\gamma}_{x},   \end{bmatrix},~ \boldsymbol{\gamma}_{x} = \begin{bmatrix}
          \cdots & \gamma_{i,x} & \cdots \end{bmatrix}. 
  \end{equation} 
\end{subequations}
In contrast to~\cite{delama2023uvio}, we have to solve for $\frac{N(N-1)}{2}$ biases $\boldsymbol{\gamma}_{x}$ originating from multiple tags or anchors.

Please note that the solution of linear least squares problem is sensitive to the noise in the reference position estimate and the distance measurement. Thus, a nonlinear refinement step is required. 

\subsubsection{Nonlinear Refinement}\label{sec:problem-initialization-nonlinear-refinement}

As proposed in~\cite{delama2023uvio}, a nonlinear refinement based on the Levenberg-Marquardt (LM) algorithm is executed on the same set of measurements for each anchor with the linear solution as initial guess, in order to minimize the following cost function 
\begin{equation}
    \min_{\vx} \frac{1}{2} || \vy - \vf(\vx) ||^{2},
\end{equation}
where $\vx = \begin{bmatrix} \reference{\vp}{\cG}{\cA_x}{} &  \boldsymbol{\gamma}_{x} &  \boldsymbol{\beta}_{x} \end{bmatrix}$ is vector to be estimated, $\vy$ are the measured outputs, and $\hat{\vy} = \vf(\vx)$ are the estimated outputs according to~\cref{eq:UWB-meas-range}. In the nonlinear refinement step, the symmetric distance-dependent bias between ranging devices $\boldsymbol{\beta}_{x}$ is estimated and initialized to $\boldsymbol{1}$. For details on the LM algorithm, we would like to refer interested reader to~\cite{delama2023uvio}.

\subsubsection{Robust Calibration}\label{sec:problem-initialization-robust-calibration}

\begin{figure}
    \centering 
    \includegraphics[trim={11 40 29 40},clip,width=0.56\columnwidth]{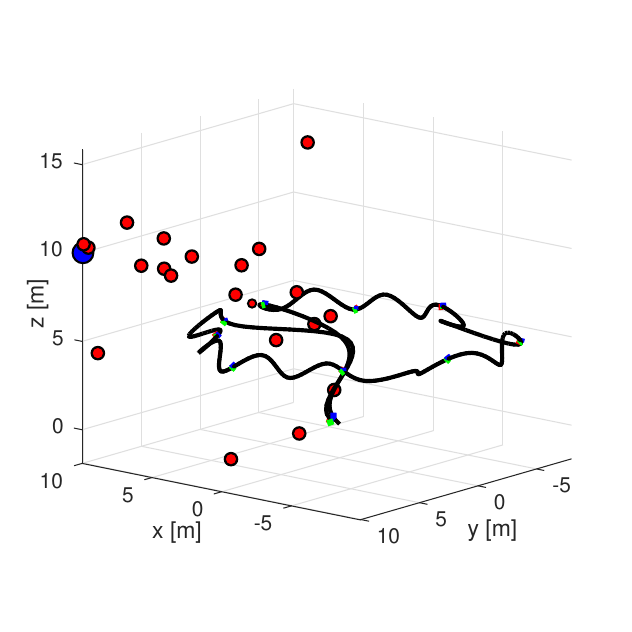}
    \includegraphics[trim={30 0 310 10},clip,width=0.42\columnwidth]{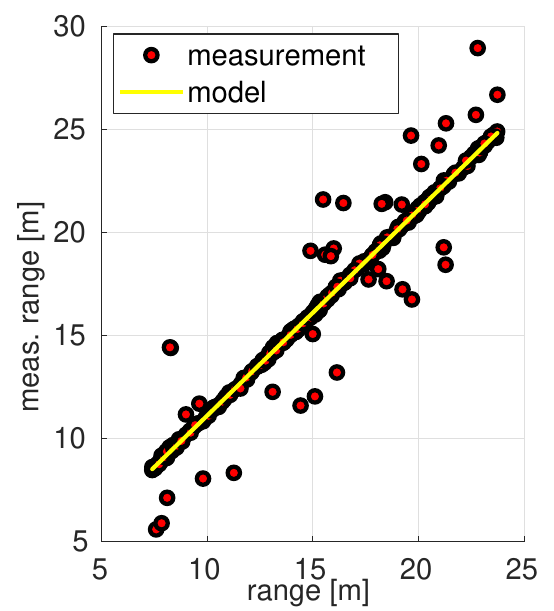}
    \caption{{Left image shows the body trajectory (black line), the stationary anchor $\reference{\vp}{\cG}{\cA}{} = [10;10;10]$ (blue circle), intermediate results (red circle) using random samples of the recorded measurements shown in the right images. The measurements contained \unit[15]{\%} outliers with a ranging and tag position standard deviation of \unit[0.1]{m}. The yellow line shows the fitted model, after removing outliers.}}
    \label{fig:robust-calibration-RANSAC}
    \vspace{-4mm}
\end{figure}

As the standard least squares problem is sensitive to outliers, we implemented an outlier detection method based on \ac{RANSAC}~\cite{fischler1981Ransac}. We assume that the number of collected measurements exceeds the number of measurements needed to solve our model. Generally, the location where the data was collected is more important than the actual sample size. In each iteration we take a random set of measurements, solve the problem in~\cref{sec:problem-initialization-double-method}, and compute a cost $|| \vy - \vf(\hat{\vx}) ||^{2}$ using the estimated model $\hat{\vx}$ and all measurements $\vy$. After a certain number of iterations, which depends on the percentage of outliers expected $\epsilon$ and the probability to obtain once a sample set containing inliers only, we search for the estimated model with the least cost. Given that model, we reject all measurements exceeding the \unit[68]{\%} of the distribution $d_{\text{thres}} = (\sigma_{d} + \sigma_{p})$, with the ranging noise $\sigma_{d}$ and tag position noise $\sigma_{p}$, and finally solve the problem using the inlier set. In~\cref{fig:robust-calibration-RANSAC}, simulation results utilizing our robust calibration method on noisy data with $\epsilon = \unit[15]{\%}$ outliers are shown.

\section{Experiments}
\label{sec:Experiments}

{For the evaluations in~\cref{sec:Evaluations}, we recorded data from a \ac{UAV} and stationary anchors in an indoor environment. The \ac{UAV} was equipped with a Pixhawk 4 flight controller and a companion computer (\ac{RPi4}), an IMU, a barometer, and two \ac{UWB} transceiver (Qorvo MDEK1001). The UAV was flown within a volume of roughly $5 \times 8 \times \unit[8]{m^3}$, covered by a motion capturing system, tracking the pose of the \ac{UAV} at \unit[60]{Hz} with sub-millimeter and sub-degree accuracy for ground truthing.
Nine stationary \ac{UWB} nodes, each consisting of a \ac{UWB} transceiver and a computer (\ac{RPi4}), were arbitrarily placed in the environment and their true positions were captured as shown in~\cref{fig:hw-setup}.
The SDS-TWR protocol~\cite{sang2018analytical} and the proposed fully meshed ranging scheme, see~\cref{fig:scheduling}, were implemented on the \ac{UWB} transceivers. For the experiments, the mesh cycle rate (\ref{eq:update_rate}) for $N=11$ nodes with $\Delta t = \unit[10]{ms}$ is $f_{\text{mesh}} = \unit[1]{Hz}$.

\begin{figure}
    \centering
    \includegraphics[width=1\linewidth]{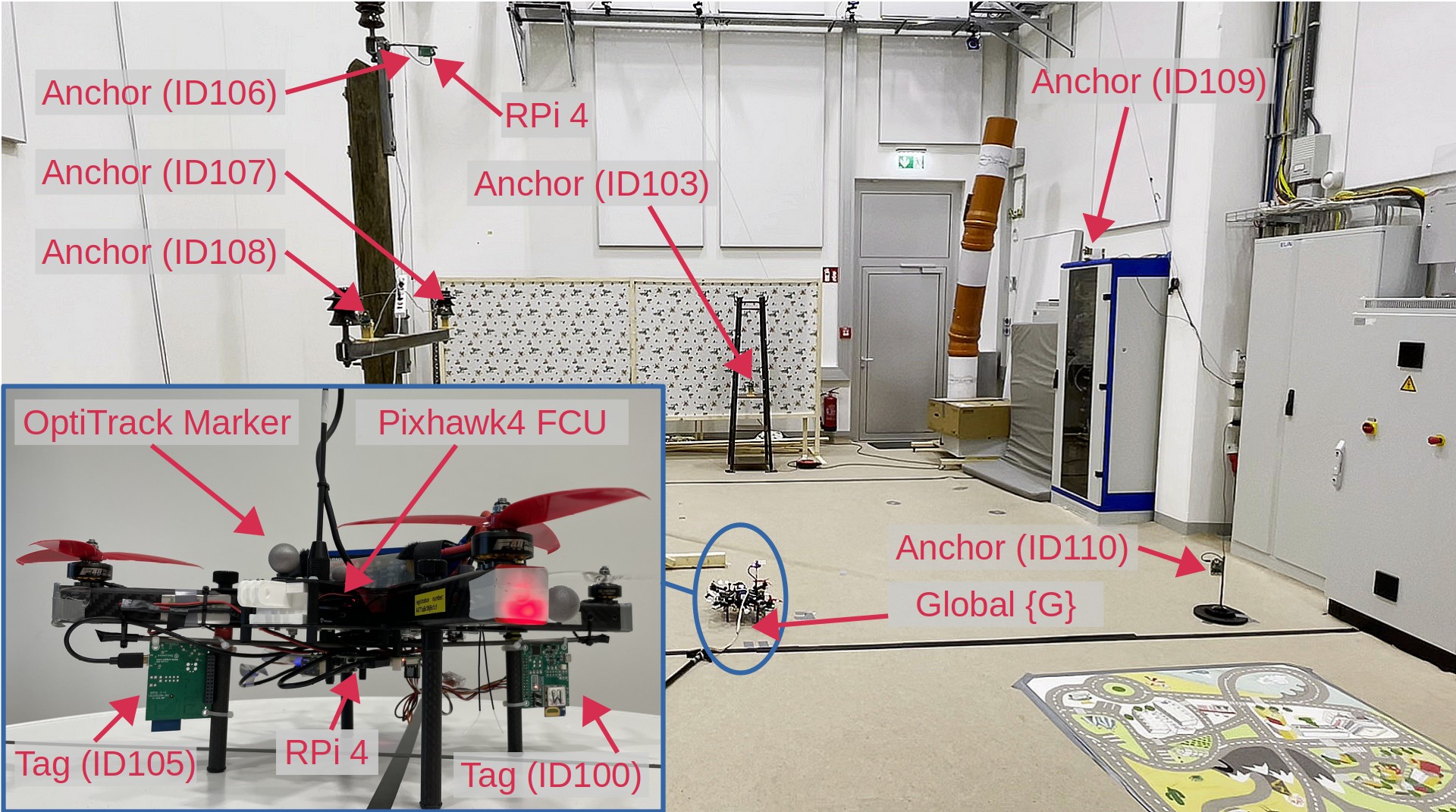}
    \caption{{Shows a subset of the deployed \ac{UWB} ranging devices in a cluttered environment, covered by a motion capture system. In total nine anchors consisting of a \ac{RPi4} companion computer and a Qorvo MDEK1001 \ac{UWB} transceiver are used. The UAV is equipped with two ranging devices, a \ac{RPi4}, a Pixhawk 4 flight controller, and reflective markers for ground truthing.}}
    \label{fig:hw-setup}
    \vspace{-4mm}
\end{figure}

The \ac{RPi4} attached to the \ac{UWB} modules receives via the UART interface raw \ac{TWR} range measurements and the unique ID of the TWR initiator and responder. A process is parsing and delegating these samples utilizing the ROS1 middleware~\cite{ROS1_Quigley_2009} over Wi-Fi 5 to subscribing nodes, e.g., the \ac{UAV}, as shown in~\cref{fig:block-diagram-hardware}. The time synchronization for accurate time stamping is performed via a network time synchronization protocol. 
 
\begin{figure}[t]
  \centering
  \includegraphics[width=0.95\columnwidth]{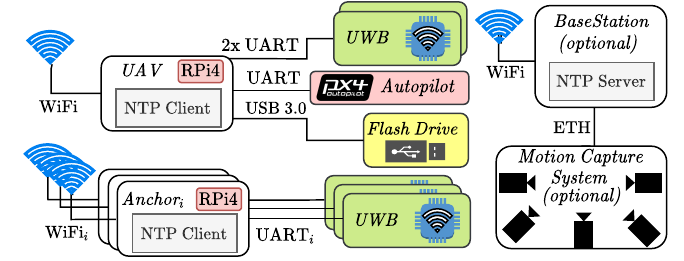}
    \caption{{Hardware components and interfaces used for the experiment, consisting of (i) a \emph{BaseStation} for mission control, time synchronization and ground truthing, (ii) a \emph{UAV} executing the mission and sensor fusion on a RPi4 utilizing two \ac{UWB} ranging devices and Pixhawk4 FCU, and (iii) a set of \emph{Anchors}, each providing local \ac{UWB} range data through the ROS1 middleware.}}
    \label{fig:block-diagram-hardware}
    \vspace{-6mm}
\end{figure}

The \textit{Mission Sequencer} of CNSFlightStack~\cite{scheiber2022cns} was used to perform autonomous waypoint following. The proposed modular estimation framework ({\tt mmsf\_ros}) and the modified \ac{UWB} anchor calibration tool originating from UVIO~\cite{delama2023uvio}, are an integral part of our software framework, as depicted in \cref{fig:block-diagram-system}. Through the ROS1 dynamic reconfigure interface, operators can execute different missions, trigger data collection, initiate calibration for new UWB anchors, and record data.

\begin{figure}[t]
    \centering
    \includegraphics[width=0.9\columnwidth]{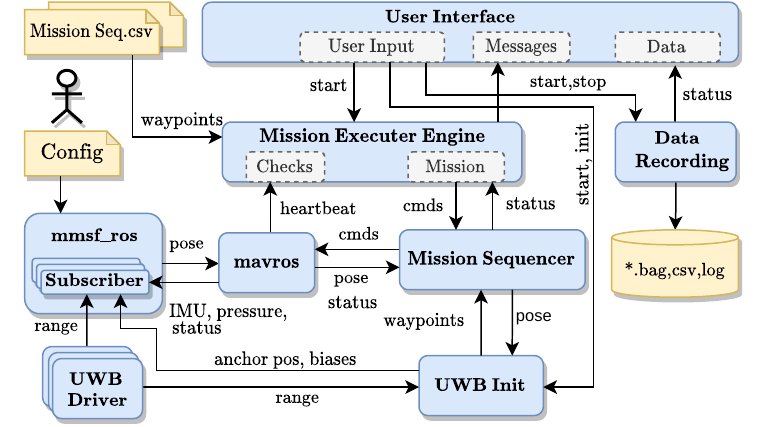}
    \caption{{The proposed system’s software components and their interaction. The sensor fusion node, \emph{mmsf\_ros}, obtains sensor data from \emph{mavros} and the \emph{UWB driver} nodes, and provides an estimated 6-DoF pose to \emph{mavros}, the \emph{Mission Sequencer}~\cite{scheiber2022cns}, and the calibration node \emph{UWB Init}. Once sufficient ranges and poses are recorded by the calibration node, the calibration is triggered and a set of anchor position and biases estimates are sent to \emph{mmsf\_ros} to instantiate new estimator instances with the provided initial beliefs.}}
    \label{fig:block-diagram-system}
    \vspace{-4mm}
\end{figure}

For the sensor fusion framework, every sensor must be assigned a unique ID greater than zero. Specifically, the tags on the \ac{UAV} are assigned IDs $100$ and $105$, while the anchors are assigned to IDs ranging from $101$ to $110$. The selection of these tag IDs results in range measurements being provided at the start and midpoint of the cycle. The IMU and the barometer are assigned to the IDs $1$ and $2$.
}

\subsection{Datasets}

During 12 autonomous flights, the flight controller was provided with the measured position of the motion capture system. The UAV took off to $2$~m, executed three upward spirals reaching $8$~m, and landed at its initial position. Anchors were randomly placed along the spiral flight path at heights ranging from $0.3$~m to $3.3$~m in an area of $5\times \unit[8]{m^2}$. Each flight lasted around $140$~s.  

\begin{figure}[t]
    \centering 
    \includegraphics[trim={0 0 0 11.2},clip,width=\columnwidth]{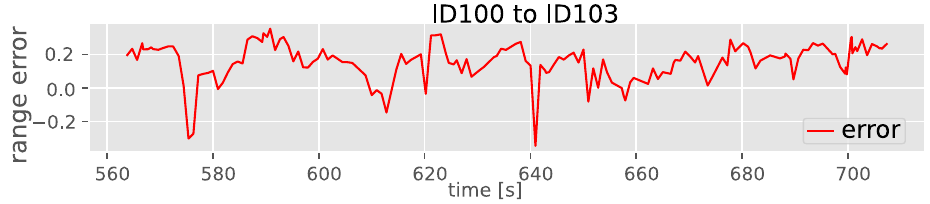}
    \includegraphics[width=\columnwidth]{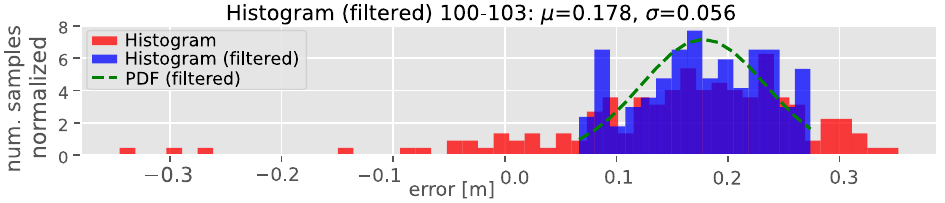} 
    \caption{{Ranging error histogram between the first tag (ID100) and the third anchor (ID103) during the first sequence of the dataset after removing significant outliers. In the bottom image, a fitted Gaussian distribution on a filtered Histogram (blue) as green dashed line is shown, to determine the constant biases $\gamma_{i,k}$ and the measurement noise $n_{d}$. The error histogram in red unveils, that the ranging errors in the dataset do not obey a Gaussian distribution.}}
    \label{fig:Eval-S1-Hist-ID100-to-ID103}
    \vspace{-4mm}
\end{figure}

To assess the measurement noise and pairwise biases of the collected fully-meshed range measurements, we developed a range evaluation package\footnote{\scriptsize \url{https://github.com/aau-cns/cnspy_ranging_evaluation}} in Python. Its computes the true distances between a set of tags moving along a given trajectory and a set of known stationary anchors at the moments the real noisy range measurement was taken by interpolating the between two closest the tag poses. Given all measured and true ranges, the error and the histogram can be calculated, and a Gaussian distribution can be fitted as shown in~\cref{fig:Eval-S1-Hist-ID100-to-ID103}. This error plot unveils, that fusing these measurements in a Bayesian estimator is generally challenging, as the error is not normally distributed.       

Since the accuracy of \ac{TOA} measurements is typically influenced by factors such as the spatial relationship between device antennas, pairwise biases, and range-based biases, we substitute the actual range measurements with the range evaluation package in 12 recorded flight datasets with true ranges, with different Gaussian noise $\sigma_{d} = \{0.1, 0.2\}$~m$^2$, and different percentage of outliers $\epsilon = \{0.0, 0.15\}$. These recorded and synthetic datasets are made available online~\cite{jung_2024_uwbdataset}.

\section{Evaluations}
\label{sec:Evaluations}

The objective of the offline evaluations is to assess the performance to the fly-by-initialization and the filter through empirical data collected in multiple runs. 

\subsection{$\Scenario_1$: Meshed Anchor Calibration Evaluation}
\label{sec:Eval-Anchor-Calib}

{In this section, we evaluate the anchor position and pairwise bias calibration approach using multiple tags (IDs: 100, 105) (see~\cref{tab:S1-anchor-calib-RANSAC}) with their known true position on 12 recorded sequences of the dataset. Remarkably, the outlier rejection is not improving the calibration result, which indicates, that the nonlinear refinement step in~\cref{sec:problem-initialization-nonlinear-refinement} suppresses the few outliers sufficiently. On average the anchor position calibration error $\reference{\tilde{\vp}}{\cG}{\cA_i}{}$ was $\unit[0.22]{m}$ in a range between $\unit[0.07]{m}$ up to $\unit[0.46]{m}$. These values can be considered as baseline for the further evaluations. Please note, that the calibration accuracy depends on the \ac{GDOP} and thus on the trajectory performed~\cite{delama2023uvio}.}   

\bgroup
\def\arraystretch{1.5}
{
\begin{table}[t]
\centering
\resizebox{\columnwidth}{!}{%
\begin{tabular}{c|cccccccccc|}
\cline{2-11}
\multicolumn{1}{l|}{}   & \multicolumn{10}{c|}{$[m]$}                                                                                                                                                                                                                                                                                                                                                                                                                                                                                                                                                                                                                                                         \\ \hline
\multicolumn{1}{|c|}{R} & \multicolumn{1}{c|}{$\referencet{\tilde{\vp}}{\cG}{\cA{1}}{}{}$} & \multicolumn{1}{c|}{$\referencet{\tilde{\vp}}{\cG}{\cA{2}}{}{}$} & \multicolumn{1}{c|}{$\referencet{\tilde{\vp}}{\cG}{\cA{3}}{}{}$} & \multicolumn{1}{c|}{$\referencet{\tilde{\vp}}{\cG}{\cA{4}}{}{}$} & \multicolumn{1}{c|}{$\referencet{\tilde{\vp}}{\cG}{\cA{6}}{}{}$} & \multicolumn{1}{c|}{$\referencet{\tilde{\vp}}{\cG}{\cA{7}}{}{}$} & \multicolumn{1}{c|}{$\referencet{\tilde{\vp}}{\cG}{\cA{8}}{}{}$} & \multicolumn{1}{c|}{$\referencet{\tilde{\vp}}{\cG}{\cA{9}}{}{}$} & \multicolumn{1}{c|}{$\referencet{\tilde{\vp}}{\cG}{\cA{10}}{}{}$} & $\referencet{\bar{\tilde{\vp}}}{\cG}{\cA}{}{}$ \\ \hline
\multicolumn{1}{|c|}{0} & \multicolumn{1}{c|}{\textbf{0.28}}                                & \multicolumn{1}{c|}{\textbf{0.46}}                                & \multicolumn{1}{c|}{0.1}                                          & \multicolumn{1}{c|}{0.15}                                         & \multicolumn{1}{c|}{\textbf{0.12}}                                & \multicolumn{1}{c|}{\textbf{0.14}}                                & \multicolumn{1}{c|}{\textbf{0.16}}                                & \multicolumn{1}{c|}{0.16}                                         & \multicolumn{1}{c|}{0.4}                                           & \textbf{0.22}                                  \\ \hline
\multicolumn{1}{|c|}{1} & \multicolumn{1}{c|}{0.37}                                         & \multicolumn{1}{c|}{\textbf{0.46}}                                & \multicolumn{1}{c|}{\textbf{0.07}}                                & \multicolumn{1}{c|}{\textbf{0.14}}                                & \multicolumn{1}{c|}{\textbf{0.12}}                                & \multicolumn{1}{c|}{0.26}                                         & \multicolumn{1}{c|}{0.17}                                         & \multicolumn{1}{c|}{\textbf{0.14}}                                & \multicolumn{1}{c|}{\textbf{0.29}}                                 & \textbf{0.22}                                  \\ \hline
\end{tabular}%
}
\caption{Anchor calibration error $\reference{\tilde{\vp}}{\cG}{\cA_i}{}$ averaged over 12 sequences of the dataset with and without RANSAC (R) using the true tag positions.}
\label{tab:S1-anchor-calib-RANSAC}
\vspace{-4mm}
\end{table}
} 

\subsection{$\Scenario_2$: Fly-by initialization evaluations on the dataset}
\label{sec:Eval-Est-Perform}

{For this offline evaluation on the 12 recorded sequences, we initiated the estimation framework with the positions, pairwise biases, and measurement noise of four known anchors (IDs: 101, 102, 103, 14) and  two tags (IDs: 100, 105), an IMU, and a barometer. The zero-velocity detection was triggered by the FCU's landing detection. The biases and \ac{UWB} ranging measurement noise was obtained \apri using the range evaluation toolbox and the true trajectory of the \ac{UAV} and known anchor positions from the first sequence.

To evaluate the anchor position initialization and self-calibration, we calibrate the remaining unknown anchors (IDs: 106-110) in two steps to demonstrate the increased computational load utilizing the \ac{DP} in the filter framework. Additionally, we compare the calibration results with and without the proposed outlier rejection method.

In the first step, range measurements are collected after the take-off from each tag (IDs: 100, 105) and \apri known anchors (IDs: 101-104) to two unknown anchors (IDs: 106, 107). At $t=\unit[80]{s}$, the calibration routine is triggered and the two new anchors' initial positions estimate $\referencet{{\vp}}{\cG}{\cA_{\{106,107\}}}{}{\star}$ with the pairwise biases between $\{100-105\}$ and $\{106,107\}$ are provided to the estimation framework through a ROS1 message. These positions and biases are used to create new anchor instances $\{\Sensor_{106}, \Sensor_{107}\}$ in the callback of the estimator and are provided to the instance handler $\InstanceHandler$. Now, range measurements between these newly established anchors can be fused, and their position estimates are refined.
The second calibration step is triggered  at $t=\unit[100]{s}$, and the remaining anchors instances $\Sensor_{\{108, 109, 110\}}$ are created and provided to the estimation framework. 
As shown in~\ref{fig:S2-execution-time}, the computational loads increases with an increasing net measurements rate according to~\cref{eq:update_rate}. It clearly illustrates the inherent scalability challenges of the estimation problem, especially when employing the centralized-equivalent \ac{DP} filter strategy.
\cref{fig:Eval-S2-Trajectory-Init} shows the true and estimated anchor constellation, and the true and estimated trajectory of the \ac{UAV} of the second sequence using the \ac{DAH} filter strategy. 
\begin{figure}[t]
  \centering 
    \includegraphics[width=0.75\columnwidth]{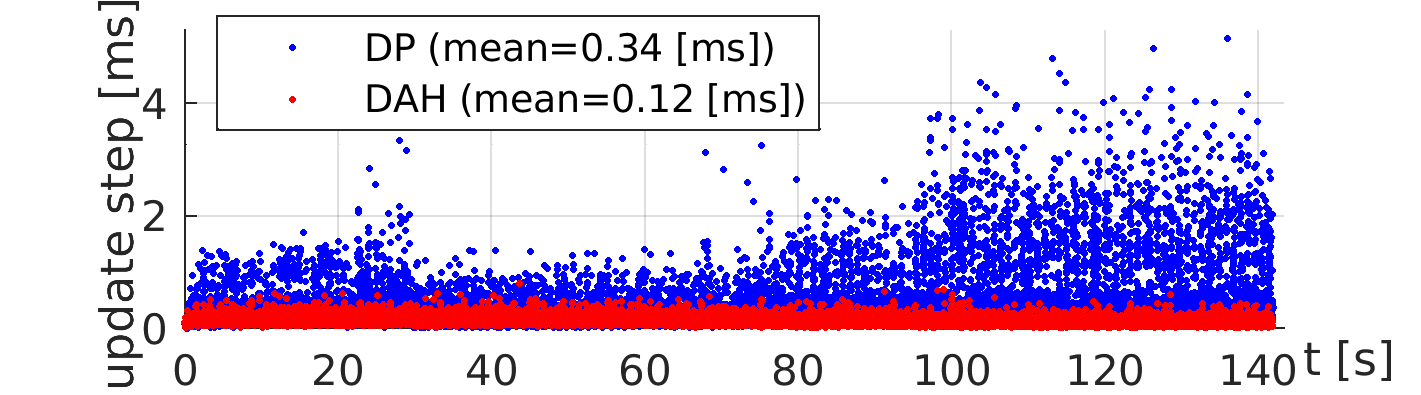}
    \caption{{Simulation $\Scenario_2$: Shows the execution time on desktop CPU for processing updates in the filter framework utilizing \ac{DP} (blue) and \ac{DAH} (red) on the second sequence. At $t=\unit[80]{s}$ two anchors are added (IDs: 106, 107) and at $t=\unit[100]{s}$ three anchors (IDs: 108, 109, 110).}} 
  \label{fig:S2-execution-time}
  \vspace{-4mm}
\end{figure}

\begin{figure}[t]
  \centering 
  \includegraphics[trim={55 50 100 80},clip,width=0.8\columnwidth]{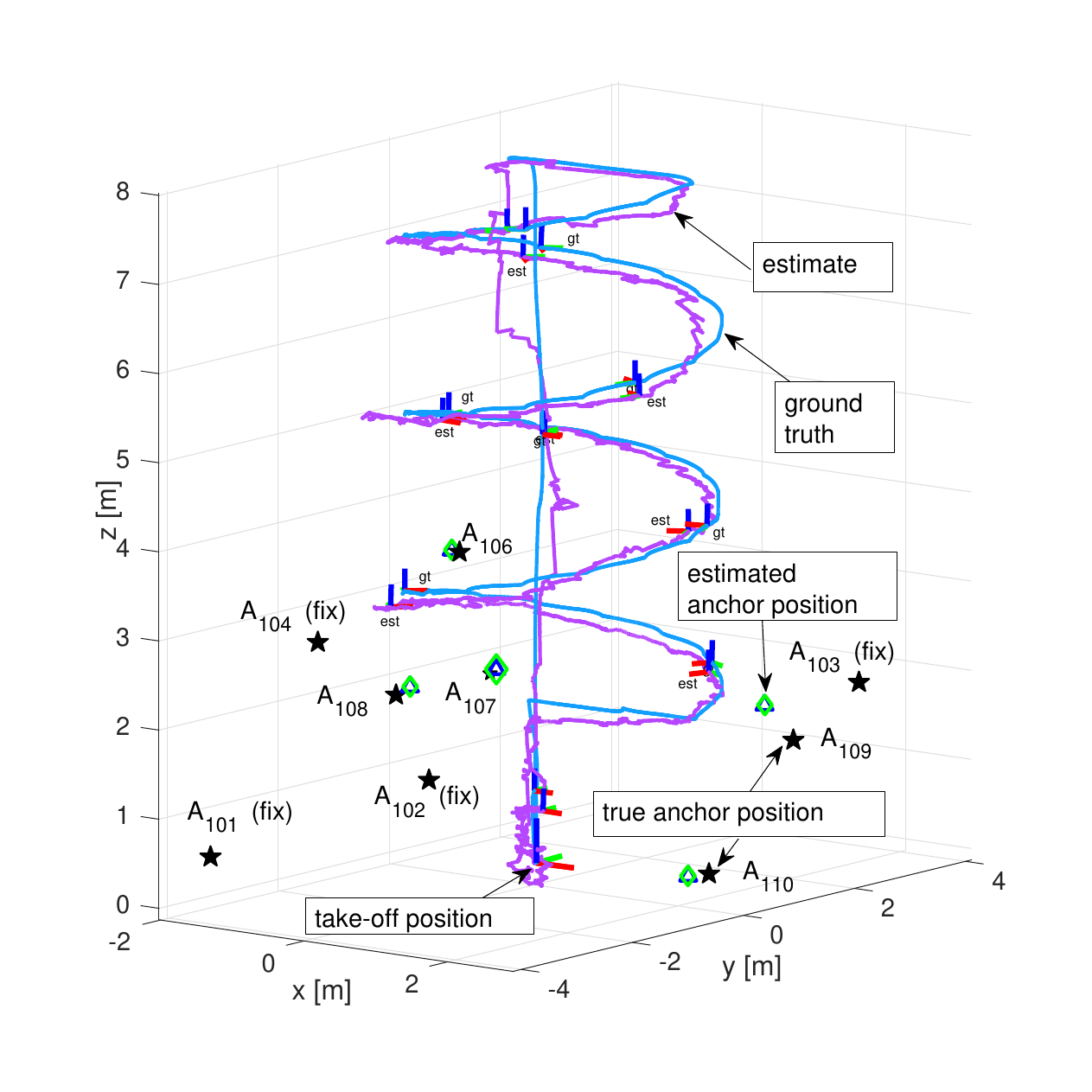}
  \caption{{Simulation $\Scenario_2$: Second sequences utilizing the DAH strategy. In blue and purple the true and estimated trajectory, respectively. Blue triangles are the initial anchor positions $\referencet{{\vp}}{\cG}{\cA}{}{\star}$, green diamonds the final estimates $\reference{\hat{\vp}}{\cG}{\cA}{}$, black stars are the true anchor positions $\reference{{\vp}}{\cG}{\cA}{}$.}}
  \label{fig:Eval-S2-Trajectory-Init}
  \vspace{-6mm}
\end{figure}

In~\cref{tab:S1-ARME-Comp}, the average \ac{RMSE} of the estimated \ac{IMU} pose and the estimated fly-by initialized anchor positions are listed. On average, the anchor position calibration with estimated tag positions is roughly \unit[0.1]{m} worse than in the previous evaluation with the true tag positions, see~\cref{tab:S1-anchor-calib-RANSAC}. The \ac{DAH} strategy is on a par with the \ac{DP} strategy, even though approximations are made during the filter update steps. 
The initial anchor estimates $\referencet{{\vp}}{\cG}{\cA}{}{\star}$ did not change significantly while being estimated in the filter framework during the remaining \unit[50]{s} or \unit[30]{s} of the trajectory. The reasons are manifold. The \ac{GDOP} of the tags decreases towards the end of the spiral leading to higher horizontal estimation errors. The initial beliefs provided by the calibration tool can be inconsistent. E.g., the \ac{NEES} of $\referencet{\tilde{\vp}}{\cG}{\cA_{7}}{}{\star}$ on the second sequence is $159.6$ and exceeding the lower \unit[99.7]{\%} confidence threshold of $13.93$. The initial covariance can be inflated by increasing the sensor noise parameter in the calibration framework.

Another interesting insight is that the outlier rejection method slightly degraded the anchor position calibration on real data, even thought the number of measurements rejected was low. E.g., during the calibration of $\referencet{\tilde{\vp}}{\cG}{\cA_{7}}{}{\star}$ on the second sequence, 4 measurements out of 715 were removed and led to error of \unit[0.35]{m}. Without outlier rejection, the error was \unit[0.2]{m}. In contrast, the outlier rejection method significantly improved the anchor calibration result on a synthetic dataset with \unit[10]{\%} outliers and $\sigma_d = \unit[0.1]{m}$, from an average error of $\unit[0.33]{m}$ to $\unit[0.23]{m}$. 
Nevertheless, a comparison with average calibration error of roughly \unit[0.25]{m} from UVIO~\cite{delama2023uvio}, which uses a modular \ac{VIO} estimation framework,  underlines the capabilities of the proposed framework.}

\bgroup
\def\arraystretch{1.5}
{
\begin{table*}[t]
\vspace{2mm}
\centering
\resizebox{1.75\columnwidth}{!}{%
\begin{tabular}{ccc|c|c|cccccccccccc|}
\cline{4-17}
                              &                               &   & $[m]$                                 & $[deg]$                               & \multicolumn{12}{c|}{$[m]$}                                                                                                                                                                                                                                                                                                                                                                                                                                                                                                                                                                                                                                                                                                                                                                                                                                  \\ \hline
\multicolumn{1}{|c|}{dataset} & \multicolumn{1}{c|}{strategy} & R & $\reference{\tilde{\vp}}{\cG}{\cI}{}$ & $\reference{\tilde{\vR}}{\cG}{\cI}{}$ & \multicolumn{1}{c|}{$\referencet{\tilde{\vp}}{\cG}{\cA_{6}}{}{\star}$} & \multicolumn{1}{c|}{$\referencet{\tilde{\vp}}{\cG}{\cA_{6}}{}{}$} & \multicolumn{1}{c|}{$\referencet{\tilde{\vp}}{\cG}{\cA_{7}}{}{\star}$} & \multicolumn{1}{c|}{$\referencet{\tilde{\vp}}{\cG}{\cA_{7}}{}{}$} & \multicolumn{1}{c|}{$\referencet{\tilde{\vp}}{\cG}{\cA_{8}}{}{\star}$} & \multicolumn{1}{c|}{$\referencet{\tilde{\vp}}{\cG}{\cA_{8}}{}{}$} & \multicolumn{1}{c|}{$\referencet{\tilde{\vp}}{\cG}{\cA_{9}}{}{\star}$} & \multicolumn{1}{c|}{$\referencet{\tilde{\vp}}{\cG}{\cA_{9}}{}{}$} & \multicolumn{1}{c|}{$\referencet{\tilde{\vp}}{\cG}{\cA_{10}}{}{\star}$} & \multicolumn{1}{c|}{$\referencet{\tilde{\vp}}{\cG}{\cA_{10}}{}{}$} & \multicolumn{1}{c|}{$\referencet{\bar{\tilde{\vp}}}{\cG}{\cA}{}{\star}$} & $\referencet{\bar{\tilde{\vp}}}{\cG}{\cA}{}{}$ \\ \hline
\multicolumn{1}{|c|}{real}    & \multicolumn{1}{c|}{DP}       & 0 & \textbf{0.2}                          & \textbf{4.02}                         & \multicolumn{1}{c|}{\textbf{0.18}}                                     & \multicolumn{1}{c|}{0.18}                                         & \multicolumn{1}{c|}{0.32}                                              & \multicolumn{1}{c|}{0.32}                                         & \multicolumn{1}{c|}{0.2}                                               & \multicolumn{1}{c|}{0.2}                                          & \multicolumn{1}{c|}{0.7}                                               & \multicolumn{1}{c|}{0.7}                                          & \multicolumn{1}{c|}{\textbf{0.35}}                                      & \multicolumn{1}{c|}{\textbf{0.35}}                                 & \multicolumn{1}{c|}{0.35}                                                & 0.35                                           \\ \hline
\multicolumn{1}{|c|}{real}    & \multicolumn{1}{c|}{DAH}      & 0 & 0.21                                  & 4.18                                  & \multicolumn{1}{c|}{\textbf{0.18}}                                     & \multicolumn{1}{c|}{0.18}                                         & \multicolumn{1}{c|}{\textbf{0.31}}                                     & \multicolumn{1}{c|}{\textbf{0.31}}                                & \multicolumn{1}{c|}{0.2}                                               & \multicolumn{1}{c|}{0.2}                                          & \multicolumn{1}{c|}{0.67}                                              & \multicolumn{1}{c|}{0.67}                                         & \multicolumn{1}{c|}{\textbf{0.35}}                                      & \multicolumn{1}{c|}{\textbf{0.35}}                                 & \multicolumn{1}{c|}{0.35}                                                & 0.34                                           \\ \hline
\multicolumn{1}{|c|}{real}    & \multicolumn{1}{c|}{DP}       & 1 & \textbf{0.2}                          & 4.06                                  & \multicolumn{1}{c|}{0.19}                                              & \multicolumn{1}{c|}{0.18}                                         & \multicolumn{1}{c|}{0.52}                                              & \multicolumn{1}{c|}{0.52}                                         & \multicolumn{1}{c|}{0.19}                                              & \multicolumn{1}{c|}{0.19}                                         & \multicolumn{1}{c|}{0.26}                                              & \multicolumn{1}{c|}{0.26}                                         & \multicolumn{1}{c|}{0.41}                                               & \multicolumn{1}{c|}{0.41}                                          & \multicolumn{1}{c|}{0.31}                                                & 0.31                                           \\ \hline
\multicolumn{1}{|c|}{real}    & \multicolumn{1}{c|}{DAH}      & 1 & \textbf{0.2}                          & 4.12                                  & \multicolumn{1}{c|}{\textbf{0.18}}                                     & \multicolumn{1}{c|}{\textbf{0.17}}                                & \multicolumn{1}{c|}{0.51}                                              & \multicolumn{1}{c|}{0.51}                                         & \multicolumn{1}{c|}{\textbf{0.17}}                                     & \multicolumn{1}{c|}{\textbf{0.17}}                                & \multicolumn{1}{c|}{\textbf{0.2}}                                      & \multicolumn{1}{c|}{\textbf{0.2}}                                 & \multicolumn{1}{c|}{0.38}                                               & \multicolumn{1}{c|}{0.38}                                          & \multicolumn{1}{c|}{\textbf{0.29}}                                       & \textbf{0.29}                                  \\ \hline \hline
\multicolumn{1}{|c|}{syn}     & \multicolumn{1}{c|}{DAH}      & 1 & \textbf{0.23}                         & \textbf{3.16}                         & \multicolumn{1}{c|}{\textbf{0.18}}                                     & \multicolumn{1}{c|}{\textbf{0.17}}                                & \multicolumn{1}{c|}{\textbf{0.25}}                                     & \multicolumn{1}{c|}{\textbf{0.25}}                                & \multicolumn{1}{c|}{\textbf{0.28}}                                     & \multicolumn{1}{c|}{\textbf{0.26}}                                & \multicolumn{1}{c|}{\textbf{0.27}}                                     & \multicolumn{1}{c|}{\textbf{0.23}}                                & \multicolumn{1}{c|}{\textbf{0.18}}                                      & \multicolumn{1}{c|}{\textbf{0.18}}                                 & \multicolumn{1}{c|}{\textbf{0.23}}                                       & \textbf{0.22}                                  \\ \hline
\multicolumn{1}{|c|}{sym}     & \multicolumn{1}{c|}{DAH}      & 0 & 0.25                                  & \textbf{3.16}                         & \multicolumn{1}{c|}{0.24}                                              & \multicolumn{1}{c|}{0.21}                                         & \multicolumn{1}{c|}{0.34}                                              & \multicolumn{1}{c|}{0.36}                                         & \multicolumn{1}{c|}{0.39}                                              & \multicolumn{1}{c|}{0.38}                                         & \multicolumn{1}{c|}{0.4}                                               & \multicolumn{1}{c|}{\textbf{0.37}}                                & \multicolumn{1}{c|}{0.25}                                               & \multicolumn{1}{c|}{0.28}                                          & \multicolumn{1}{c|}{0.33}                                                & 0.32                                           \\ \hline
\end{tabular}%
}
\caption{Simulation $\Scenario_2$: Shows the estimation error averaged over 12 sequences of the \ac{IMU} pose and the fly-by initialized anchors, utilizing different fusion strategies, with and without outlier rejection (R), and different datasets. The synthetic (syn) dataset contained unbiased noisy ($\sigma_d = 0.1$) range measurements with $\epsilon = \unit[10]{\%}$ outliers.}
\label{tab:S1-ARME-Comp}
\vspace{-5mm}
\end{table*}
} 

\subsection{$\Experiment_1$: Closed loop fly-by initialization}
\label{sec:Experiments-Real}

{In this section, we evaluate the calibration and estimation performance during an autonomous closed-loop flight. Therefore, the filter framework and calibration framework were executed in real-time on the companion computer (\ac{RPi4}) of the \ac{UAV} and provided an estimated pose to the flight controller. 
To enhance the overall robustness of the system, we used in total five fixed anchors. As part of this adjustment, we adjusted the set of waypoints to lower the height of the spiral trajectory to \unit[2.2]{m}, with alternating levels between \unit[1]{m} and \unit[2.2]{m}, as depicted in \cref{fig:E1-Trajectory}.

\begin{figure}[t]
  \centering 
  \includegraphics[trim={42 48 55 45},clip,width=0.9\columnwidth]{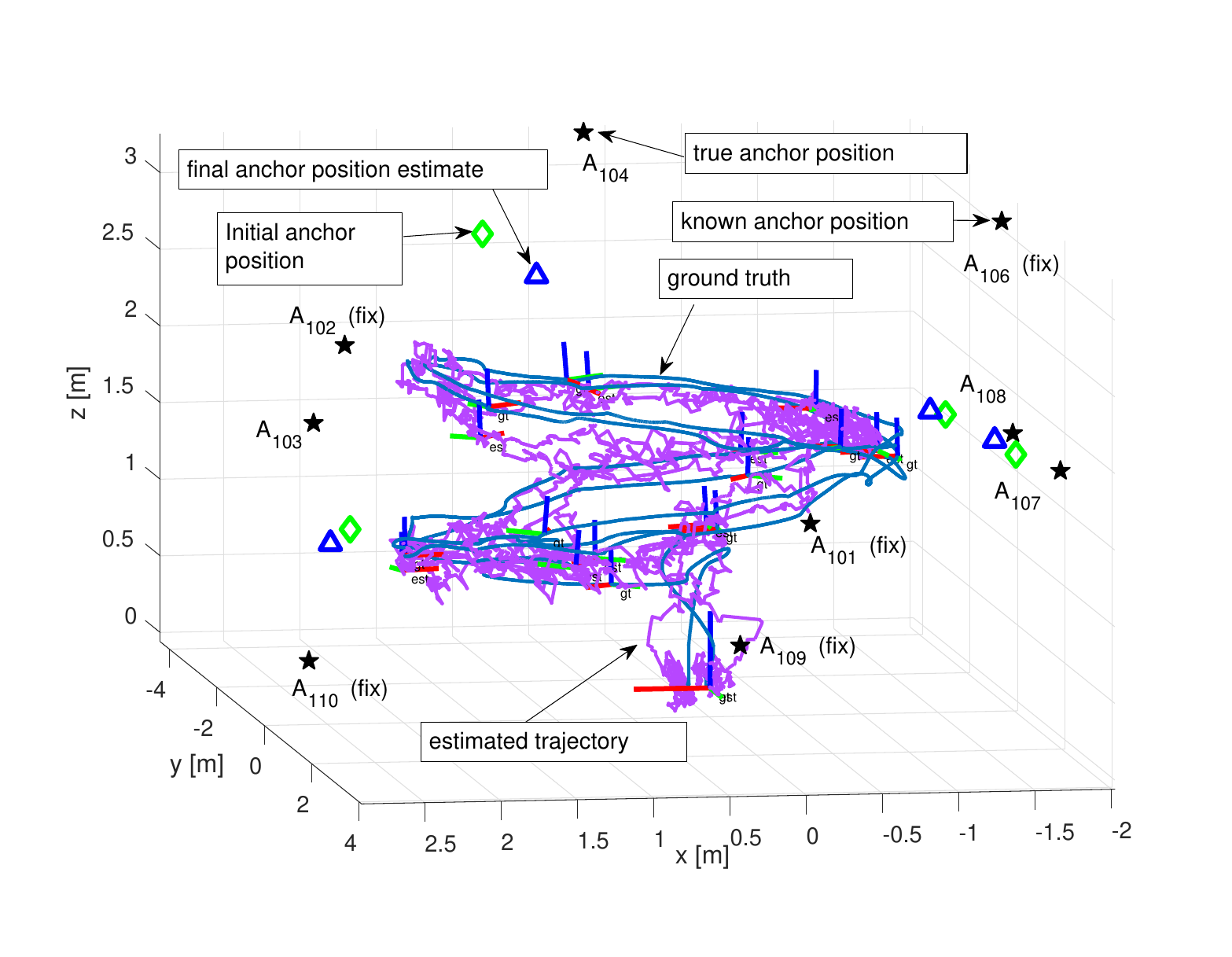}
  \caption{Experiment $\Experiment_1$: Closed loop flight utilizing the \ac{DAH} strategy with five fixed anchors $\reference{\vp}{\cG}{\cA_{\{3,4,7,8\}}}{}$. In blue and purple the true and estimated trajectory, respectively. Blue triangles are the initial anchor positions $\referencet{{\vp}}{\cG}{\cA}{}{\star}$, green diamonds the final estimates $\reference{\hat{\vp}}{\cG}{\cA}{}$, black stars are the true anchor positions $\reference{{\vp}}{\cG}{\cA}{}$. }
  \label{fig:E1-Trajectory}
  \vspace{-4mm}
\end{figure}

However, during some trials, the estimator experienced a huge orientation error about the z-axis, as indicated in \cref{fig:E1-Est-Err}, while the position error remain bounded. One possible cause is the sudden gyroscope bias drift in the measurements provided by the flight controller and is subject to future investigations. 
%
\begin{figure}[t]
  \centering 
  \includegraphics[trim={30 0 30 0},clip,width=0.7\columnwidth]{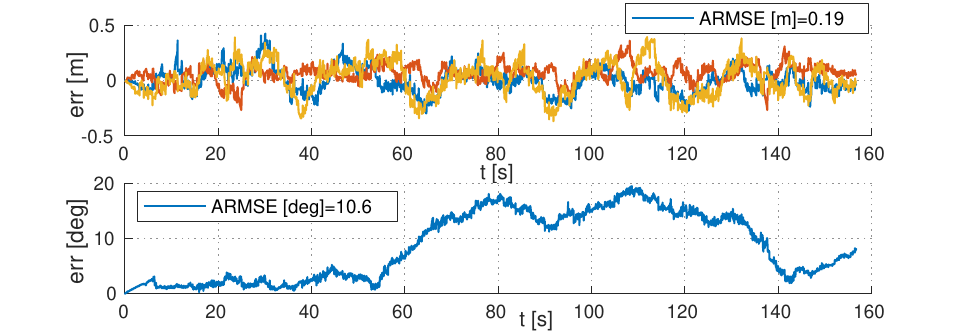}
  \caption{Experiment $\Experiment_1$: Pose error plot of the closed loop flight utilizing \ac{DAH} strategy. Top image shows the position error of the x-, y-, z-axis held in blue, red, and orange, respectively. Bottom image show the orientation error. }
  \label{fig:E1-Est-Err}
  \vspace{-4mm}
\end{figure}
Nevertheless, the system was still able to perform a coarse anchor initialization, followed by an online self-calibration of the anchor position estimates, as detailed in \cref{tab:E1-error}. Compared to the offline evaluations of the dataset, the initialization error was significantly higher and originates from erroneous position estimates (max. error was \unit[0.55]{m}).
%
%
\begin{table}[t]
    \centering
    \resizebox{0.8\columnwidth}{!}{%
    
\begin{tabular}{|c|c|cccccc|}
\hline
$[m]$                                 & $[deg]$                               & \multicolumn{6}{c|}{$[m]$}                                                                                                                                                                                                                                                                                                                                                                                \\ \hline
$\reference{\tilde{\vp}}{\cG}{\cI}{}$ & $\reference{\tilde{\vR}}{\cG}{\cI}{}$ & \multicolumn{1}{c|}{$\referencet{\tilde{\vp}}{\cG}{\cA{3}}{}{}$} & \multicolumn{1}{c|}{$\referencet{\tilde{\vp}}{\cG}{\cA{4}}{}{}$} & \multicolumn{1}{c|}{$\referencet{\tilde{\vp}}{\cG}{\cA{7}}{}{}$} & \multicolumn{1}{c|}{$\referencet{\tilde{\vp}}{\cG}{\cA{8}}{}{}$} & \multicolumn{1}{c|}{$\referencet{\bar{\tilde{\vp}}}{\cG}{\cA}{}{\star}$} & $\referencet{\bar{\tilde{\vp}}}{\cG}{\cA}{}{}$ \\ \hline
0.19                                  & 10.6                                  & \multicolumn{1}{c|}{0.57}                                         & \multicolumn{1}{c|}{0.75}                                         & \multicolumn{1}{c|}{0.37}                                         & \multicolumn{1}{c|}{0.33}                                         & \multicolumn{1}{c|}{0.5}                                                 & 0.5                                            \\ \hline
\end{tabular}%
    }
    \caption{Experiment $\Experiment_1$: Shows the \ac{RMSE} of the estimated \ac{IMU} pose and the fly-by initialized anchor positions $\referencet{\tilde{\vp}}{\cG}{\cA}{}{}$ for the anchors $\cA_{\{3,4,7,8\}}$ using the DAH strategy. $\referencet{\bar{\tilde{\vp}}}{\cG}{\cA}{}{\star}$ and $\referencet{\bar{\tilde{\vp}}}{\cG}{\cA}{}{}$ are the average error of the initial and refined anchor estimate.}
    \label{tab:E1-error}
 \vspace{-5mm}
\end{table}
%
The average execution times of filter on the \ac{RPi4} were $\unit[1.1]{ms}$ for update steps using DP and $\unit[0.28]{ms}$ using DAH. The \ac{IMU} propagation steps took on average $\unit[2.1]{ms}$ using DP and $\unit[0.56]{ms}$ using DAH. It's important to note that out-of-order measurements are handled during the propagation steps, which resulted occasionally in slightly longer execution times.

In this initial evaluation, we successfully demonstrated the real-time capability and true modularity of the proposed estimation framework. Potential improvements could be achieved by adjusting the geometry of the anchor positions or by following optimal waypoints after a coarse initialization~\cite{delama2023uvio}. }
\vspace{-3mm}
\section{Conclusion}\label{sec:conclusion}

{In this paper, a truly modular sensor fusion framework, that is capable of processing any-to-any \ac{UWB} ranging measurements efficiently, is presented. This is particularly interesting for fully-meshed \ac{UWB} ranging and allows treating single \ac{UWB} devices (tag or anchors) as logical instances that can be added and removed during the mission. A meshed ranging protocol based on SDS-TWR was implemented, and a dataset was recorded in an indoor environment. The dataset and a range evaluation toolbox are made available online. To fully exploit meshed ranging in the anchor calibration procedure and the filter framework, adaptions to a state-of-the-art calibration framework were made. The effectiveness of this framework is validated through offline evaluations on the dataset and real-world experiments, by running it in real-time on a UAV. 
Future work will focus on an extension to multi-agent scenarios.}

\section{Acknowledgment}\label{sec:acknowledgement}

The authors would like to thank Giulio Delama and Alessandro Fornasier for assistance in utilizing their \ac{UWB} initialization package~\cite{delama2023uvio} and sharing their experiences.

\balance
\bibliographystyle{bib/IEEEtranBST/IEEEtran}
\bibliography{bib/references_clean}

\end{document}